\documentclass[conference]{IEEEtran}
\IEEEoverridecommandlockouts
\usepackage{cite}
\usepackage{booktabs}
\usepackage{amsmath,amssymb,amsfonts}
\usepackage{algorithmic}
\usepackage{graphicx}
\usepackage{textcomp}
\usepackage{xcolor}
\usepackage{multirow}
\def\BibTeX{{\rm B\kern-.05em{\sc i\kern-.025em b}\kern-.08em
    T\kern-.1667em\lower.7ex\hbox{E}\kern-.125emX}}
\begin{document}

\title{GT-PCQA: Geometry-Texture Decoupled Point Cloud Quality Assessment with MLLM}


\author{
\begin{tabular}{c}
Guohua Zhang$^{1}$ \quad 
Jian Jin$^{2}$ \quad 
Meiqin Liu$^{1*}$\thanks{Corresponding author: mqliu@bjtu.edu.cn} \quad 
Chao Yao$^{3}$ \quad 
Weisi Lin$^{2}$ \quad
Yao Zhao$^{1}$ \\
\\
$^{1}$Beijing Jiaotong University \\
$^{2}$Nanyang Technological University \\
$^{3}$University of Science and Technology Beijing \\
\end{tabular}
}
\maketitle

\begin{abstract}
With the rapid advancement of Multi-modal Large Language Models (MLLMs), MLLM-based Image Quality Assessment (IQA) methods have shown promising generalization.
However, directly extending these MLLM-based IQA methods to PCQA remains challenging.
On the one hand, existing PCQA datasets are limited in scale, which hinders stable and effective instruction tuning of MLLMs.
On the other hand, due to large-scale image-text pretraining, MLLMs tend to rely on texture-dominant reasoning and are insufficiently sensitive to geometric structural degradations that are critical for PCQA. 
To address these gaps, we propose a novel MLLM-based no-reference PCQA framework, termed GT-PCQA, which is built upon two key strategies.
First, to enable stable and effective instruction tuning under scarce PCQA supervision, a 2D–3D joint training strategy is proposed.
This strategy formulates PCQA as a relative quality comparison problem to unify large-scale IQA datasets with limited PCQA datasets. It incorporates a parameter-efficient Low-Rank Adaptation (LoRA) scheme to support instruction tuning.
Second, a geometry-texture decoupling strategy is presented, which integrates a dual-prompt mechanism with an alternating optimization scheme to mitigate the inherent texture-dominant bias of pre-trained MLLMs, while enhancing sensitivity to geometric structural degradations.
Extensive experiments demonstrate that GT-PCQA achieves competitive performance and exhibits strong generalization.
\end{abstract}

\section{Introduction}
\label{sec:intro}
Point clouds, as 3D representations of objects or scenes, are widely used in applications such as Virtual Reality (VR) \cite{zhang2022openpointcloud}, Augmented Reality (AR) \cite{guo2020deep}, 3D modeling \cite{mekuria2016design}, and autonomous driving \cite{li2025image}. However, point clouds are inevitably degraded during acquisition, processing, storage, and transmission \cite{gao2023opendmc,11419157,yang2026scaling,chen2025mugsqa}. Therefore, accurate Point Cloud Quality Assessment (PCQA) metrics are crucial. 

PCQA metrics are commonly classified into Full-Reference (FR), Reduced-Reference (RR), and No-Reference (NR) methods. Early efforts primarily focused on FR metrics, which calculate distortions based on complete reference information, ranging from geometric distances like MSE-p2pl and HD-p2pl \cite{mse-p2pl} to color-based PSNR-yuv \cite{psnr-yuv} and structural similarity-based descriptors such as GraphSIM \cite{graphsim} and PointSSIM \cite{pointssim}. Since obtaining the original reference point cloud is often difficult in real-world scenarios, this also makes NR-PCQA more critical and challenging. 

Recently, NR-PCQA has experienced significant improvement through the use of advanced Deep Neural Networks (DNNs) \cite{gao2023opendmc}. These approaches have evolved from traditional hand-crafted statistical models such as 3D-NSS \cite{3dnss} to more sophisticated deep architectures, including ResSCNN \cite{resscnn}, PQA-net \cite{liu2021pqa},  and the MM-PCQA \cite{mm-pcqa}, which integrates multi-modal features. 
Despite their promising performance on individual benchmarks, these methods are typically optimized on specific datasets, causing the learned quality representations to be closely tied to dataset-specific distortion characteristics. As a result, their performance often degrades under distribution shifts between training and testing data, leading to limited cross-dataset generalization.

To improve the generalization of NR-PCQA, various advanced training techniques have been explored, like the domain adaptation \cite{yang2022no,lu2024styleam,sun2025mfcqa,zhang2026qd}. For the traditional image quality assessment, Multi-modal Large Language Models (MLLMs) have demonstrated promising cross-dataset generalization for Image Quality Assessment (IQA) tasks. Among MLLM-based IQA methods \cite{li2025perceptual,wu2023q}, Compare2Score~\cite{zhu2024adaptive}, which formulates IQA as a relative quality comparison, enables MLLMs to leverage supervision from multiple datasets, providing a scalable training paradigm for robust quality assessment.

However, directly extending these MLLM-based IQA methods to PCQA remains challenging.
On the one hand, existing PCQA datasets are limited in scale, which hinders stable and effective instruction tuning of MLLMs.
On the other hand, due to large-scale image-text pretraining, MLLMs tend to rely on texture-dominant reasoning and are insufficiently sensitive to geometric structural degradations that are critical for PCQA. 




To address these challenges, we propose an MLLM-based GT-PCQA that includes two key strategies.
First, to enable stable and effective instruction tuning of MLLMs under scarce PCQA supervision, a 2D–3D joint training strategy is proposed. Specifically, PCQA is formulated as a relative quality comparison problem, which not only serves as a unified bridge for jointly leveraging large-scale IQA datasets and limited PCQA datasets, but also naturally aligns with instruction–response learning. This comparative formulation facilitates the construction of scalable and reusable supervision signals, allowing the model to learn from both image- and point cloud-based quality annotations in a unified framework. To further support stable and efficient instruction tuning under limited PCQA supervision, we integrate a parameter-efficient Low-Rank Adaptation (LoRA) \cite{hu2021lora} scheme. It constrains task-specific updates to a low-rank subspace, preserving the general visual–language capabilities of the pre-trained MLLM while enabling the effective learning of distortion-aware representations. 
Second, to mitigate the inherent texture-dominant bias of pre-trained MLLMs, while enhancing sensitivity to geometric structural degradations, a geometry–texture decoupling strategy is presented. This strategy employs a dual-prompt mechanism combined with an alternating optimization scheme, explicitly separating the learning of geometry-aware and texture-aware representations.
Specially, the dual-prompt mechanism applies geometry-aware and texture-aware prompts to multi-view point clouds and images, respectively, guiding the model to attend to geometric structural degradations without suppressing texture cues, thereby preserving their texture awareness and enhancing sensitivity to geometric structural degradations.
Furthermore, under this alternating optimization scheme, geometry-focused and texture-focused optimization steps are strictly alternated during fine-tuning, thereby mitigating the inherent texture-dominant bias of pre-trained MLLMs.
The contributions can be summarized as follows:
\begin{itemize}
\item We propose the GT-PCQA, which formulates PCQA as a unified comparative evaluation task, effectively unifying image- and point cloud-based quality assessment under a relative comparison paradigm.
\item We propose a 2D-3D joint training strategy that leverages relative quality comparison to integrate large-scale IQA datasets with limited PCQA datasets and uses a parameter-efficient LoRA scheme, enabling stable and effective instruction tuning.
\item We propose a geometry-texture decoupling strategy, consisting of a dual-prompt mechanism and an alternating optimization scheme, which mitigates the inherent texture-dominant bias of pre-trained MLLMs, while enhancing sensitivity to geometric structural degradations.
\end{itemize}


\section{Proposed Method}
\begin{figure*}[t]
\centering
\includegraphics[width=\linewidth]{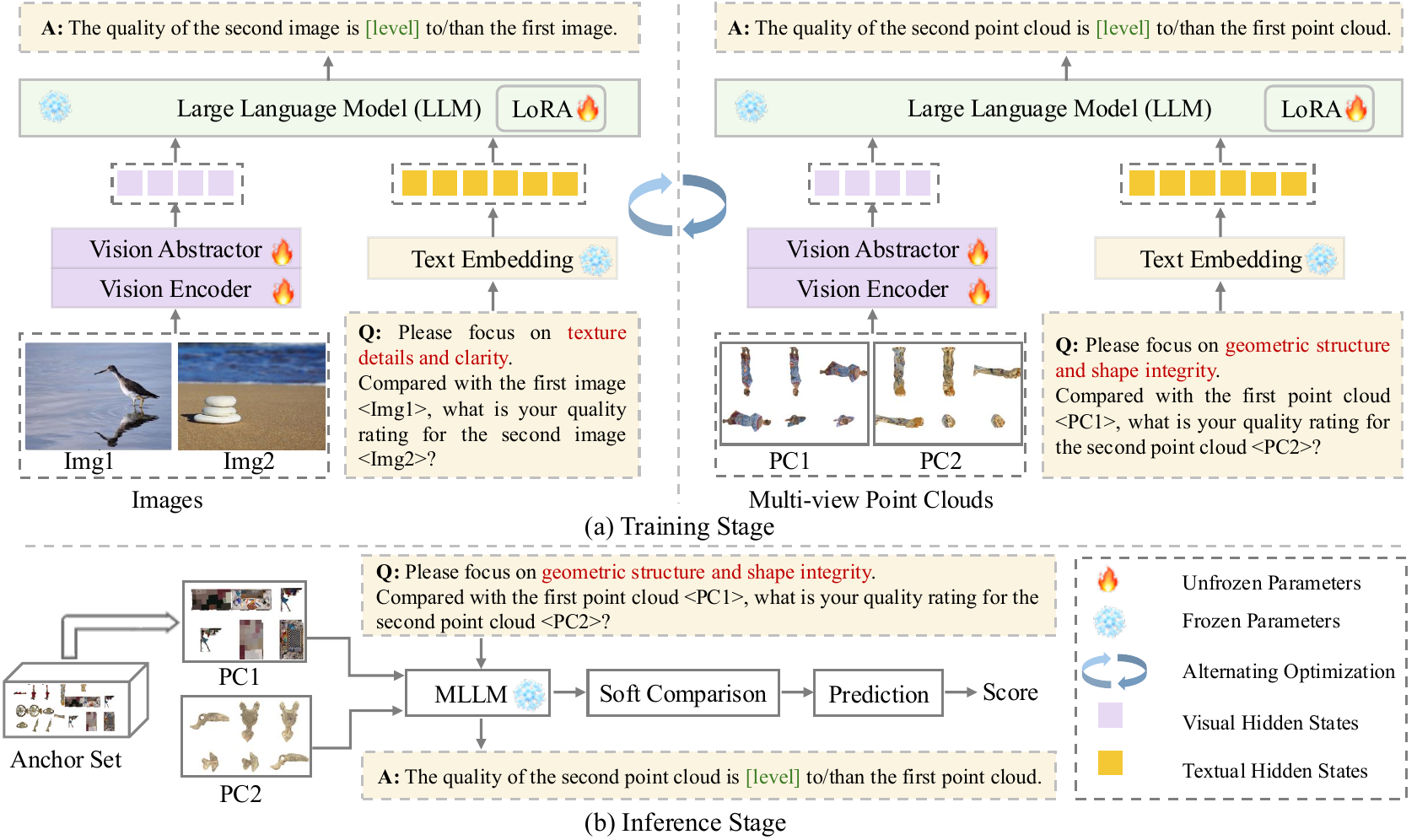}
\caption{ Architecture of the proposed GT-PCQA. (a) During the training stage, the model alternates between image pairs and multi-view point cloud pairs. Visual inputs are encoded by the vision encoder and abstracted into compact representations, while attribute-specific text prompts (\emph{e.g.,} texture-aware or geometry-aware) are embedded into textual representations. The aligned multimodal features are then fed into a LoRA-adapted LLM to perform relative quality comparison, enabling stable and effective instruction tuning under heterogeneous 2D–3D supervision.
(b) During the inference stage, the trained MLLM is fully frozen, and each test point cloud is evaluated against an anchor set via soft comparison, ultimately predicting the final quality score.
}
\vspace{-1.5em}
\label{fig:1}
\end{figure*}

An overview of the proposed GT-PCQA framework is illustrated in Fig.~\ref{fig:1}, and the details are introduced in the following subsections.

\subsection{2D–3D Joint Training Strategy}
To enable stable and effective instruction tuning of Multi-modal Large Language Models (MLLMs) under scarce PCQA supervision, a 2D–3D joint training strategy is proposed that formulates PCQA as a relative quality comparison problem, serving as a unified bridge to jointly leverage large-scale IQA datasets and limited PCQA datasets. This comparative formulation naturally aligns with instruction–response learning and enables the construction of scalable and reusable supervision signals for instruction tuning.

Since subjective evaluation protocols vary across datasets and lead to inconsistent perceptual scales, we adopt a pairwise quality comparison mechanism for images and multi-view point clouds. This design enables the construction of large-scale and reusable instruction–response pairs. Following empirical criteria~\cite{parametrichandbook}, quality differences are discretized into five levels: inferior, worse, similar, better, and superior.
Following common practice in subjective quality modeling \cite{zhu2024adaptive}, we assume that the rating uncertainties of two samples are independent, which allows the quality difference between them to be normalized by their aggregated uncertainty.

Formally, we randomly sample $n_k$ image or multi-view point clouds pairs
$\{(x_k^{(i)}, x_k^{(j)})\}_{i,j=1}^{n_k}$ from each dataset. For each sampled pair $\left(x^{(i)}, x^{(j)}\right)$, the corresponding normalized quality level $L_{ij}$, which represents the relative quality of sample $j$ with respect to sample $i$, is formulated as:

\begin{equation}
\label{eq:quality_level}
L_{ij}=
\begin{cases}
\text{inferior}, & \text{if } z_{ij} > 2 \\
\text{worse},    & \text{if } 1 < z_{ij} \leq 2 \\
\text{similar},  & \text{if } |z_{ij}| \leq 1 \\
\text{better},   & \text{if } -2 < z_{ij} \leq -1 \\
\text{superior}, & \text{if } z_{ij} \leq -2 
\end{cases},
\end{equation}
\noindent where $z_{ij}$ denotes the standardized quality difference between samples $i$ and $j$, formulated as:
\begin{equation}
\label{eq:z_score}
z_{ij} = \frac{q^{(i)} - q^{(j)}}{\sqrt{(\sigma^{(i)})^2 + (\sigma^{(j)})^2}} ,
\end{equation}
\noindent where $q^{(i)}$ and $q^{(j)}$ represent the Mean Opinion Score (MOS) of samples
$i$ and $j$, respectively, while $\sigma^{(i)}$ and $\sigma^{(j)}$ denote their corresponding rating standard deviations.

To further ensure that instruction tuning remains stable and effective under scarce PCQA supervision, the joint training process is implemented using a parameter-efficient adaptation scheme based on Low-Rank Adaptation (LoRA), which restricts task-specific updates to a low-rank subspace while preserving the general visual–language capabilities of the pre-trained MLLM \cite{hu2021lora}.


Specifically, the visual encoder and visual abstractor are fully fine-tuned to capture distortion-sensitive representations from both images and multi-view point clouds, enabling the model to perceive fine-grained geometric artifacts and bridge the substantial domain gap between natural images and point cloud projections.
In contrast, the large language model is adapted via lightweight LoRA modules inserted into the query and value projection matrices, while all other parameters remain frozen.

From a mechanism perspective, quality assessment within MLLMs is primarily governed by attention-based comparison and evidence aggregation \cite{vaswani2017attention}.
Updating the query projections modulates how visual and linguistic tokens attend to distortion-related cues, while adapting the value projections controls how quality-relevant evidence is propagated through attention layers \cite{vaswani2017attention}.
Therefore, restricting adaptation to the query–value subspace enables effective recalibration of quality-aware reasoning while preserving the general semantic and syntactic representations of the language model, which is consistent with the design principle of LoRA \cite{hu2021lora}.

Overall, the proposed 2D–3D joint training strategy reformulates PCQA as a relative quality comparison task, enabling large-scale IQA data to be reused as unified and reusable instruction-level supervision.
Together with parameter-efficient LoRA-based adaptation, it enables stable and effective instruction tuning under scarce PCQA supervision, aligning quality-aware reasoning across image and point cloud domains.
\subsection{Geometry–Texture Decoupling Strategy}
To mitigate the inherent texture-dominant bias of pre-trained MLLMs, while enhancing sensitivity to geometric structural degradations, a geometry–texture decoupling strategy is presented that integrates a dual-prompt mechanism with an alternating optimization scheme to decouple geometry-aware and texture-aware learning explicitly.

\subsubsection{Dual-Prompt Mechanism}
The dual-prompt mechanism applies geometry-aware and texture-aware prompts to multi-view point clouds and images, respectively, guiding the model to attend to geometric structures without suppressing texture cues, thereby preserving their texture awareness and enhancing sensitivity to geometric structural degradations.

There are two prompts inluding a texture prompt $P_T$ for images and a geometry prompt $P_G$ for multi-view point clouds. $P_T$ guides attention to fine-grained surface details and color, while $P_G$ emphasizes spatial structures and shapes. During training, visual features are concatenated with the corresponding prompt and fed into the LLM, steering it to focus on quality-relevant cues.

The instruction-response formats for image and point cloud pairs are defined as follows:

\begin{quote}
\textbf{Texture prompt:} Please focus on texture details and clarity. Compared with the first image \textless Img1\textgreater, what is your quality rating for the second image \textless Img2\textgreater?

\textbf{Response:} The quality of the second image is [level] to/than the first image.
\end{quote}

\begin{quote}
\textbf{Geometry prompt: }Please focus on geometric structure and shape integrity. Compared with the first point cloud \textless PC1\textgreater, what is your quality rating for the second point cloud \textless PC2\textgreater?

\textbf{Response:} The quality of the second point cloud is [level] to/than the first point cloud.
\end{quote}

\subsubsection{Alternating Optimization Scheme}
To mitigate the inherent texture-dominant bias of pre-trained MLLMs, an alternating optimization scheme is proposed.

Specifically, the optimization process alternates between geometry-aware and texture-aware updates during fine-tuning. At each optimization step $t$, the model is guided by either geometry-oriented or texture-oriented supervision, ensuring that geometric cues are periodically reinforced rather than being suppressed by the stronger texture bias inherited from pre-trained MLLMs.

The training loss $\mathcal{L}_{ce}^{(t)}$ at step $t$ is formulated as:
\begin{equation}
\mathcal{L}_{ce}^{(t)} =
\begin{cases}
\mathbb{E}_{x \sim \mathcal{D}_T} \big[ \mathrm{CE}(x) \big], & \text{if } t \text{ is even} \\
\mathbb{E}_{x \sim \mathcal{D}_G} \big[ \mathrm{CE}(x) \big], & \text{if } t \text{ is odd}
\end{cases},
\end{equation}
where $\mathcal{D}_T$ and $\mathcal{D}_G$ denote the texture-oriented and geometry-oriented data distributions, respectively.
The cross-entropy loss $\mathrm{CE}(x)$ is formulated as:
\begin{equation}
\mathrm{CE}(x) = - \sum_{c=1}^{C} y_c \log p_c(x),
\end{equation}
where $p_c(x)$ is the predicted probability of the $c$-th relative quality level, $C$ is the total number of levels, and $y_c$ is the corresponding ground-truth level label.

By separating geometry- and texture-driven updates, our mechanism prevents bias toward texture cues and ensures stable learning of geometry-aware quality features, which is crucial for robust PCQA.

After training, the MLLM is frozen and used in an inference stage, where the learned multimodal feature embeddings are utilized within an MLLM-based soft comparison ~\cite{zhu2024adaptive} to evaluate each test point cloud by comparing it against an anchor set for quality prediction.
Following this framework, an anchor set $\mathcal{A}$ is constructed from the SJTU-PCQA dataset to ensure reliable pairwise comparisons.
Specifically, the dataset is partitioned into $\beta$ quality intervals, and one anchor object is selected from each interval based on rating consistency.
The anchor set $\mathcal{A}$ is formulated as:
\begin{equation}
\mathcal{A} = \bigcup_{k=1}^{\beta} a_k,
\end{equation}
where $a_k$ denotes the anchor object corresponding to the $k$-th quality interval, which is selected by minimizing the variance of subjective scores, formulated as:
\begin{equation}
a_k = \underset{x \in \mathcal{D}_k}{\arg \min} \, \sigma^2(x),
\end{equation}
\noindent where $\mathcal{D}_k$ represents the $k$-th quality interval of the SJTU-PCQA dataset, and $\sigma^2(x)$ denotes the variance of MOS scores for multi-view point cloud $x$.

Then, following the soft comparison protocol, the pairwise comparison outcomes are aggregated into a probability matrix $P$, which captures the relative quality ordering between the test sample and the anchor set. Based on this matrix, the quality prediction score $\hat{q}$ of the test sample is inferred by solving a posterior-driven optimization problem:
\begin{equation}
\hat{q} = \arg \max_q \ \mathcal{F}(q; P),
\end{equation}
\noindent where $q$ denotes the latent continuous quality variable, and $\mathcal{F}(q; P)$ denotes an objective function that encourages consistency between the estimated quality score and the observed soft comparison relations encoded in $P$, following common practices in probabilistic ranking and quality assessment~\cite{kristi2011analyze}.

\section{Experiments}
\subsection{Experimental Setup}
\subsubsection{Dataset and Evaluation Metrics}
Our GT-PCQA is trained under a joint 2D–3D setting using seven datasets in total, including the SJTU-PCQA ~\cite{yang2020predicting} dataset for PCQA supervision and six large-scale Image Quality Assessment (IQA) datasets (LIVE~\cite{live}, KADID-10k~\cite{lin2019kadid}, CLIVE~\cite{clive}, KonIQ-10k~\cite{koniq}, BID~\cite{bid}, and CSIQ~\cite{csiq}) to support stable and effective instruction tuning with LoRA.
\textbf{Since SJTU-PCQA is the only dataset providing standard deviation annotations required for our training objective}, it is used for both training and in-dataset evaluation. To assess generalization, the trained model is directly evaluated on the unseen PCQA dataset WPC~\cite{su2019perceptual} without any fine-tuning.
We adopt Pearson’s Linear Correlation Coefficient (PLCC), Spearman Rank Order Correlation Coefficient (SROCC), Kendall Rank Order Correlation Coefficient (KROCC), and Root Mean Square Error (RMSE) as metrics. Better performance is indicated by higher PLCC, SROCC, and KROCC, and lower RMSE.
\subsubsection{Implementation Details}
Our GT-PCQA is built upon the mPLUG-Owl2 framework \cite{ye2024mplug}. equipped with a CLIP-ViT-L vision encoder \cite{radford2021learning}, a six-layer Q-Former visual abstractor, and an LLaMA-2-7B \cite{touvron2023llama} as the LLM. We train the model on 210k images and multi-view point clouds pairs using the cross-entropy over predicted logits, with a batch size of 64 for 3 epochs. We adopt AdamW \cite{loshchilov2017decoupled} as the optimizer. The initial learning rate is set to 2e-5 and decays gradually using the cosine decay strategy. 
To enable parameter-efficient adaptation, LoRA adapters are inserted into the attention projection layers of the LLM. The rank is set to $r=128$, which is selected based on ablation studies, while the scaling factor is empirically set to $\alpha=256$ to match the chosen rank. All other training settings are kept consistent across different datasets. Training is conducted on three RTX 3090 GPUs, and each epoch takes approximately 5.1 hours to complete, while inference can be performed on a single RTX 3090 GPU.
Furthermore, to obtain the anchor set, we divide the training set of the SJTU-PCQA \cite{yang2020predicting} into five ($\beta = 5$) quality intervals based on their MOSs and Standard deviations, from which we select one representative anchor multi-view point cloud.
\subsection{Performance Evaluation}
As shown in Tab.~\ref{tab:1}, we compare GT-PCQA with representative full-reference (FR) and no-reference (NR) PCQA methods on the SJTU-PCQA dataset.
Among NR approaches, GT-PCQA achieves competitive performance, with an SROCC of 0.8953 and a PLCC of 0.8883, substantially outperforming several established deep learning baselines such as Compare2Score~\cite{zhu2024adaptive} (by 29.1\% in SROCC) and IT-PCQA~\cite{yang2022no} (by 40.7\% in SROCC). In addition, GT-PCQA significantly reduces the RMSE to 0.8629, corresponding to a 53.2\% error reduction compared to Compare2Score \cite{zhu2024adaptive}.

It is worth noting that MM-PCQA~\cite{mm-pcqa} achieves higher in-domain accuracy on SJTU-PCQA, largely due to its fully supervised training paradigm and dataset-specific optimization, which can lead to overfitting on the training distribution. In contrast, GT-PCQA emphasizes cross-dataset generalization by leveraging relative quality comparison and reusable instruction-level supervision, along with parameter-efficient instruction tuning. As shown in the cross-dataset evaluation (Sec.~\ref{sec:1}), GT-PCQA demonstrates substantially better generalization when applied to unseen PCQA datasets, reflecting a more favorable trade-off between in-domain performance and robustness across diverse point cloud distributions.
\begin{table}[htbp]
    \centering
   \caption{Performance comparison on SJTU-PCQA dataset. The best and second-best NR-PCQA results are highlighted in \textbf{bold} and \textit{italic}, respectively.}
    \small 
    \label{tab:1}
    \resizebox{\linewidth}{!}{ 
    \begin{tabular}{c|c|cccc}
    \toprule
    Type & Methods & SROCC$\uparrow$ & PLCC$\uparrow$ & KROCC$\uparrow$ & RMSE$\downarrow$ \\
    \midrule
    \multirow{5}{*}{FR} 
    & MSE-p2pl \cite{mse-p2pl} & 0.6277 & 0.5940 & 0.4825 & 2.2815 \\
    & HD-p2pl \cite{mse-p2pl} & 0.6441 & 0.6874 & 0.4565 & 2.1255 \\
    & PSNR-yuv \cite{psnr-yuv} & 0.7950 & 0.8170 & 0.6196 & 1.3151 \\
    & GraphSIM \cite{graphsim} & 0.8783 & 0.8449 & 0.6947 & 1.0321 \\
    & PointSSIM \cite{pointssim} & 0.6867 & 0.7136 & 0.4964 & 1.7001 \\
    \midrule
    \multirow{7}{*}{NR} 
    & ResSCNN \cite{resscnn} & 0.8600 & 0.8100 & - & - \\
    & PQA-net \cite{liu2021pqa} & 0.8372 & 0.8586 & 0.6304 & 1.0719 \\
    & 3D-NSS \cite{3dnss} & 0.7144 & 0.7382 & 0.5174 & 1.7686 \\
    & IT-PCQA \cite{yang2022no} & 0.6361 & 0.6934  & 0.4932 & 1.6240 \\
    & MM-PCQA \cite{mm-pcqa} & \textbf{0.9103} & \textbf{0.9226} & \textbf{0.7838} & \textbf{0.7716} \\
    & Compare2Score \cite{zhu2024adaptive} & 0.6933 & 0.7926  & 0.5516 & 1.8436 \\
    & GT-PCQA (Ours) & \textit{0.8953} & \textit{0.8883} & \textit{0.7382} & \textit{0.8629} \\
    \bottomrule
    \end{tabular}
    }
\end{table}
    
    
\subsection{Ablation Study}
We conduct comprehensive ablation studies to evaluate the contributions of key components in GT-PCQA and the impact of the LoRA rank. Component-level results for the Dual-Prompt (DP), Alternating Optimization (AO), and LoRA are reported in Tab.~\ref{tab:2}, while the effect of different LoRA ranks is presented in Tab.~\ref{tab:3}.

\textbf{Impact of the Dual-Prompting mechanism (DP). }Removing DP leads to a significant drop in SROCC and PLCC, indicating that DP helps the model balance geometric and texture learning, thereby preserving texture awareness while enhancing sensitivity to geometric structural degradations. As geometric structural degradations are a key factor in PCQA, DP plays a crucial role in quality discrimination.

\textbf{Impact of the Alternating Optimization scheme (AO).} Removing AO caused slight but consistent drops across all metrics, indicating that the alternating optimization mechanism is crucial for training stability. It prevents over-reliance on texture cues and consistently enhances sensitivity to geometric structures across objectives, improving overall generalization.

\textbf{Impact of the LoRA.} Removing LoRA results in the largest performance degradation, with consistently lower PLCC and KROCC and higher RMSE. This demonstrates that directly fine-tuning the language model under limited PCQA supervision leads to unstable instruction optimization. In contrast, the lightweight LoRA adaptation enables stable and effective instruction tuning of the MLLM by constraining language-side updates, while allowing the visual branch to be fully optimized for learning distortion-sensitive representations, thereby improving overall prediction accuracy.

\begin{table}[htbp]
    \centering
    \caption{Ablation studies on DP, AO, and LoRA, where ``w/o'' denotes removing the corresponding component; in particular, ``w/o LoRA'' indicates full fine-tuning of all LLM parameters.}
    \small 

    \begin{tabular}{c|cccc}
    \toprule
    Model  & SROCC$\uparrow$ & PLCC$\uparrow$ & KROCC$\uparrow$ & RMSE$\downarrow$ \\
    \midrule
    w/o DP   & 0.8045 & 0.8321 & 0.6887 & 1.2415\\
    w/o LoRA  & 0.8031 & 0.7845 & 0.5818 & 1.7629\\
    w/o AO   & 0.8345 & 0.8451 & 0.6921 & 1.2314\\
    GT-PCQA(Ours) &  \textbf{0.8953} & \textbf{0.8883} & \textbf{0.7382} & \textbf{0.8629}\\
    \bottomrule
    \end{tabular}
    \label{tab:2}
\end{table}
\textbf{Impact of LoRA Rank.}
We further analyze the effect of the LoRA rank on performance and efficiency.
As shown in Tab.~\ref{tab:3}, the ablation results reveal a clear trade-off between model capacity and generalization. Increasing the LoRA rank from $r=32$ to $r=128$ consistently improves performance, indicating that a moderate rank is necessary to effectively support the Dual-Prompt mechanism and model decoupled geometry–texture representations. However, further increasing the rank to $r=256$ results in a noticeable performance degradation, which we attribute to overfitting under limited PCQA supervision. Excessive capacity weakens the geometry-aware inductive bias introduced by the alternating optimization and gradually biases the model toward texture-dominant reasoning. Consequently, $r=128$ achieves the best balance between capacity and generalization and is used in all experiments.
\begin{table}[htbp]
    \centering
    \caption{Ablation study on LoRA rank $r$ using the SJTU-PCQA dataset. Bold indicates the best performance.}
     \small 
    \label{tab:3}
    \begin{tabular}{c|cccc}
    \toprule
    Rank ($r$) & SROCC$\uparrow$ & PLCC$\uparrow$ & KROCC$\uparrow$ & RMSE$\downarrow$ \\
    \midrule
    32  & 0.8375 & 0.8411 & 0.6515 & 3.1118 \\
    64  & 0.8562 & 0.8556 & 0.6840 & 2.8394 \\
    128 & \textbf{0.8953} & \textbf{0.8883} & \textbf{0.7382} & \textbf{0.8629} \\ 
    256 & 0.8601 & 0.8241 & 0.6638 & 1.6082 \\
    \bottomrule
    \end{tabular}
\end{table}
\begin{table}[htbp]
    \centering
    \caption{Cross-dataset evaluation from SJTU-PCQA to WPC. GT-PCQA is trained on SJTU-PCQA, and no fine-tuning is performed on WPC. Bold indicates the best performance.}
     \small 
    \begin{tabular}{c|cccc}
    \toprule
    Methods  & SROCC$\uparrow$ & PLCC$\uparrow$ & KROCC$\uparrow$ & RMSE$\downarrow$ \\
    \midrule
    MM-PCQA \cite{mm-pcqa}& 0.4988 & 0.4354 & 0.2929 & 2.2103\\
    Compare2Score \cite{zhu2024adaptive}   & 0.5157 & 0.4589 & 0.3829 & 2.0103\\
    GT-PCQA(Ours) & \textbf{0.6708} & \textbf{0.6339} & \textbf{0.4579} & \textbf{1.3672}\\
    \bottomrule
    \end{tabular}
    \label{tab:4}
\end{table}

\subsection{Cross-Dataset Validation}
\label{sec:1}
Cross-dataset generalization performance of different methods. Models are trained on SJTU-PCQA, with auxiliary IQA datasets used during training to support stable instruction tuning with LoRA. Evaluation is performed directly on the unseen WPC dataset without any fine-tuning. The performance of competitive methods, including MM-PCQA \cite{mm-pcqa} and Compare2Score \cite{zhu2024adaptive}, is reported for comparison. As shown in Tab.~\ref{tab:4}, several observations can be made: i) GT-PCQA achieves the best performance across all metrics, demonstrating strong cross-dataset generalization. ii) Despite substantial differences in distortion types, point cloud density, and content characteristics between SJTU-PCQA and WPC, GT-PCQA maintains robust performance. This indicates that the model captures intrinsic, distortion-aware geometric quality cues, and highlights the advantage of our geometry- and texture-aware dual-prompt mechanism combined with the proposed training strategy. Further cross-dataset evaluation on AGIQA-3K~\cite{agiqa} is provided in the supplemental material, demonstrating the robust generalization of GT-PCQA across diverse datasets.


\section{Conclusion}
In summary, we propose GT-PCQA, a novel MLLM-based NR-PCQA framework. To enable stable and effective instruction tuning under limited PCQA supervision, we propose a 2D–3D joint training strategy that formulates PCQA as a relative quality comparison problem, unifying large-scale IQA datasets with scarce PCQA data, and leveraging a parameter-efficient LoRA scheme to support instruction tuning. To address the inherent texture-dominant bias of pre-trained MLLMs while preserving texture awareness, we design a geometry–texture decoupling strategy that integrates a dual-prompt mechanism with alternating optimization to explicitly separate geometry- and texture-aware learning. Extensive experiments on multiple PCQA datasets demonstrate that GT-PCQA achieves highly competitive performance with strong cross-dataset generalization, strengthening the perceptual ability of MLLMs for point cloud quality assessment.

\bibliographystyle{IEEEbib}
\bibliography{main}
\end{document}